# On Effectively Predicting Autism Spectrum Disorder Using an Ensemble of Classifiers

Bhekipho Twala

Digital Transformation Portfolio, Tshwane University of Technology, Private Bag x680, Pretoria 0001, South Africa

E: twalab@tut.ac.za

Eamon Molloy

Waterford Institute of Technology, School of Science & Computing, Waterford, Ireland

E. emolloy@wie.ie

An ensemble of classifiers combines several single classifiers to deliver a final prediction or classification decision. An increasingly provoking question is whether such systems can outperform the single best classifier. If so, what form of an ensemble of classifiers (also known as multiple classifier learning systems or multiple classifiers) yields the most significant benefits in the size or diversity of the ensemble itself? Given that the tests used to detect autism traits are time-consuming and costly, developing a system that will provide the best outcome and measurement of autism spectrum disorder (ASD) has never been critical. In this paper, several single and later multiple classifiers learning systems are evaluated in terms of their ability to predict and identify factors that influence or contribute to ASD for early screening purposes. A dataset of behavioural data and robot-enhanced therapy of 3,000 sessions and 300 hours, recorded from 61 children are utilised for this task. Simulation results show the superior predictive performance of multiple classifier learning systems (especially those with three classifiers per ensemble) compared to individual classifiers, with bagging and boosting achieving excellent results. It also appears that social communication gestures remain the critical contributing factor to the ASD problem among children.

1. **INTRODUCTION**

World Autism Awareness month is celebrated in April worldwide by people and organisations alike. This entire month is dedicated to raising awareness, sharing understanding, and shedding light on a global and South African health crisis, which parents have been battling for years. Autism is a developmental disorder that affects interaction and communication while Autism Spectrum Disorder (ASD) is a developmental disability that affects social interaction, communication and learning skills (where the spectrum reflects a wide range of symptoms that the child can present). Abnormalities characterise themselves in social interactions and patterns of communication and a restricted, stereotyped, repetitive repertoire of interests and activities (WHO, 2018). However, the diagnosis of autism is tricky since there are no tests conducted. Its symptoms typically appear in the first two years of a child's life, developing in a specific period. One diagnosis proposal has been to look at a child's behaviour and development. One South African group, Quest School (Sowetan LIVE, 2022) has proposed that children be diagnosed when they are young, not at five years of age as is the norm.

We can not overemphasise the impact of the devastation of a child with autism. Such destruction includes a child facing different learning abilities, concentration, and mental health problems such as excessive worry, anxiety, depression, avoidance behaviours, meltdowns, and shutdowns. Recently, there has been an explosion in autism cases worldwide, and they have been increasing at an alarming rate. The World Health Organisation (WHO, 2017; Wolff and Piven, 2020; Hong *et al.*, 2020) have argued that 1 out of every 160 children has ASD. While a certain percentage of people with this disorder have been shown to live independently others would require life-long care and support.

Autism traits are difficult to trace due to tests and diagnoses requiring significant time and cost. Thus, developing early detection systems for autism could be of great help. This early identification will help with proper medication for patients early, thus preventing the patient's condition from deteriorating further. It would help reduce long-term costs associated with delayed diagnosis.

The topic of ASD early diagnosis and testing has been of interest to researchers for decades. There has been a lot of research work in the spheres of machine learning and statistical pattern recognition, where communities have discussed how one could combine models or model predictions (Leroy *et al.*, 2006). Furthermore, many research works in these communities have shown that an ensemble learning of classifiers is an effective technique for improving predictive accuracy (with the variance reduction benefit contributing to that). Ensemble learning of classifiers' development and successful fielding have significantly lagged in bio-medical and health science research activities, yet it has been prominent in other fields. A central concern of all these applications is the need to increase the predictive accuracy of early autism diagnosis and test decisions. Due to the nature of autism and its impact on societies, an improvement in predictive diagnosis accuracy or even a fraction of a per cent translates into significant future savings in time, costs, and even deaths (Buescher *et al.*, 2014; Soul and Spence, 2020). Furthermore, the economic effect of ASD on individuals with the disorder, their families, and society as a whole has been poorly understood and has not been updated in light of recent ASD prediction and detection findings. This enormous effect on families warrants better prediction and detection methods.

Several other studies for predicting autism traits in an individual have been carried out by the autism research and data (science) analytics community using several machine learning and statistical techniques. These include screening detection, identification, classification and prediction of autism traits in an individual. For screening detection, alternating and functional decision trees (Wall *et al.*, 2012; Kosmicki *et al.*, 2015), support vector machines (Bone *et al.*, 2016) and "red flags" (Allison *et al.*, 2016) have been used while support vector machines have been used for both detection and identification (Heinsfeld *et al.*, 2018). . Kosmicki *et al.* (2015) investigated logistic model trees, logistic regression for detecting non-ASD against ASD among children. To predict ASD traits, a support vector machine, naïve Bayes classifier, and the random forest has been further applied by Bekorom *et al.*, 2017. Deep learning and neural networks have been used by Heinsfeld *et al.* (2018) to predict ASD patients using imaging of the brain with a follow-up research work on classification and hemodynamic fluctuations by Xu *et al.* (2019).

From the review, it is evident that all the researchers did not focus on improving the accuracy of the machine learning models, henceforth, ASD predictive accuracy. Yet, proactive corrective actions can be taken well in advance if the prediction and detection of ASD are even more accurate. A slight increase in predictive accuracy will have a positive impact on foreseeing ASD in toddlers.

The basic idea behind ensemble learning is to train multiple classifier learning systems (MCLS) to achieve the same objective and then combine their predictions. There are different ways ensembles can be developed and the resulting output combined to classify new instances. The popular approaches to creating ensembles include changing the cases used for training through techniques such as bagging (Breiman, 1996), boosting (Freund and Schapire, 1996), stacking (Wolpert, 1992), changing the features used in training (Ho, 1995), and introducing randomness in the classifier itself (Dietterich, 2000).

This research work proposes autism diagnosis tests and detection models using an ensemble of classifiers approach to effectively help predict autism traits of a child aged three-year-old or more. Such an ensemble approach is mostly used to overcome precariousness in predictions and to also enhance the accuracy and efficiency of the predictions. The systems architecture and resampling processes are also considered. In other words, this research work focuses on predicting efficiency in identifying ASD traits among a group of children.

To this end, the first significant contribution of the paper is the proposal of using five single classifier learning systems (SCLS) to accurately predict and detect ASD for early screening purposes, and further help robots understand when an autistic child needs help. To find out if it would be worthwhile to overcome the limitations of SCLS and their inability to handle more complex tracking situations with high accuracies, the next contribution is the proposal that MCLS be utilised to deal with the ASD prediction problem. For the MCLS to increase the performance over individual models, the unique models must be accurate individually and they need to be sufficiently diverse. In other words, the tools need to be sufficiently different from each other in terms of which errors were made. For this reason, all possible combinations of the number of classifiers per ensemble are explored in this paper (i.e., from two classifiers per ensemble to five classifiers). Then, we identify the best-ranked features that impact ASD.

The rest of the paper is organised as follows: Section 2 gives a background on single classifier systems used for autism spectrum disorder prediction. Then multiple classifier learning systems are examined from the intelligibility viewpoint to improve autism spectrum disorder predictive accuracy (Section 3). Section 4 presents the study methodology in the form of an experimental set-up and results drawn from a DREAM dataset, supporting a data-driven study of autism spectrum disorder and robot-enhanced therapy. Finally, the paper is concluded with critical research findings and remarks in Section 5.

## 2. SINGLE CLASSIFIER SYSTEMS

There are several approaches to learning a single classifier. However, only five base methods of classifier construction are considered for this paper (i.e. a mixture of regression and tree-based, nets, instance-based and Bayesian related). These are Artificial Neural Network (ANN), Decision Tree (DT), *k*-Nearest Neighbour (*k*-NN), Logistic Discrimination (LgD) and the Naïve Bayes Classifier (NBC). A brief description of the five classifiers and their use for classification or prediction tasks is now given.

### A. Logistic discrimination

Logistic discrimination (LgD) is a supervised learning classification algorithm used to predict the probability of the target variable (for example, a class). It was initially developed by Cox (1966) and later modified by Day and Kerridge (1967). LgD is related to logistical regression due to the dichotomous nature of the dependent variable. In other words, the dependent variable can only take two possible values (either 0 for non-detection of ASD or 1 for the detection of ASD). For LgD, the probability density functions for the classes are not modelled like most supervised learning classification methods but rather the ratios between them (i.e., Lg D is partially parametric).

An unknown instance is a new element classified using a cut-off point score where the error rate is lowest for the cut-off point = 0.5 (Rumelhart *et al.*, 1986). The slope of the cumulative logistic probability function is steepest $\pi_i = 0.5$ $\pi_i \geq 0.5$ $\pi_i < 0.5$ at which the halfway point [i.e., the logit function transforms continuous values to the range (0, 1)], which is necessary since probabilities must be between 0 and 1. The LgD approach can be generalised to more than two classes (also called multinomial logit models). Multinomial Logit Models (MLMs) are derived similarly to the LgD models. For more details about MLMs, Hosmer and Lameshow (1989) referred to the interested reader.

### B. *k-Nearest Neighbour*

The *k*-nearest neighbour (*k*-NN) or instance-based learning approach is one of the most venerable and easy-to-implement machine learning algorithms for supervised and sometimes unsupervised learning (Aha *et al*., 1991). *k*-NN can solve both classification and regression (prediction) problems by assuming that similar things exist near each other. Thus, the *k*-NN hinges on this assumption being true enough for the algorithm to be valid. Essentially, *k*-NN works by assigning the classification (or regression prediction) of the nearest set of previously classified (predicted) occurrences to an unknown instance. The memory is the storage for the entire training set. To classify a new example, a distance measure (such as Hamming, Cosine similarity, Chebychev, Euclidean, Manhattan or Minkowski) is computed between the trained and unknown instances. For this paper, the Cosine similarity distance measure is used. Each stored training and the unknown instance are assigned the class of that nearest neighbouring instance. These nearest neighbours are first computed, and then the new example is given the most frequent class among the *k* neighbours. To select the value of *k* that is right for your data, the algorithm is run several times with different values of *k*. It reduces the number of errors encountered while maintaining the algorithms' ability to make predictions when given data not seen before accurately. For the paper, the process of supervised learning will be focused on.

### C. *Artificial Neural Networks*

Like most state-of-the-art classification methods, neural networks or artificial neural networks (ANNs) are non-parametric (i.e., no assumptions about the data are made, as is the case with models such as linear regression). Instead, ANNs are computational models inspired by an animal's nervous system. These are represented by connections (layers) between many simple computing processors or elements ("neurons"). They have been used for various classification and regression problems in economics, forensics, and pattern recognition. The ANN is trained by supplying it with many numerical observations or the patterns to be trained (input data pattern) whose corresponding classifications (desired output) is known. The final sum-of-squares error (SSE) over the validation data for the network is calculated when training the network. This SSE value is then used to select the optimum number of hidden nodes resulting in a trained neural network.

A new unknown instance is carried out by sending its attribute values to the network's input nodes, where weights are applied to those values. Finally, the values of the output unit activations are computed. The weights and biases can be optimised by running the network multiple times. Its most significant output unit activation determines the classification of the new instance.

### D. Decision Trees

A decision tree (DT) classifier is a supervised machine learning algorithm used for regression and classification tasks. It starts with a single node (subsequently, a series of decisions) and branches into possible outcomes, giving it a tree-like diagram. When training a DT, the best attribute is selected (using the information gain measure) from the total attributes list of the data for the root node, internal node and leaf or terminal nodes. A DT classifier is simple to understand, interpret and visualise.

According to Safavian and Landgrebe (1991), a DT classifier has four primary objectives. These are 1) Classifying the training sample correctly as much as possible; 2) Generalising beyond the training sample so that unseen samples could be classified with high accuracy; 3) To quickly updating the DT as more training samples become available (which is similar to incremental learning), and 4) Having a simple DT structure as possible. Despite the DT classifier strengths, Objective 1) is highly debatable and, to some extent, conflicts with objective 2). Also, not all DT classifiers are concerned with objective 3). DTs are non-parametric and valuable to represent the logic embodied in software routines.

A DT takes as input a case or example described by a set of attribute values and outputs a Boolean multi-valued "decision," making them easy to build automated predictive models. For this paper, the Boolean case is considered. Classifying an unknown instance is easy once the tree has been constructed. Starting from the root node of the DT and applying certain test conditions would eventually lead you to a leaf node with a class label associated with it. The class label associated with the leaf or terminal node is assigned to the instance.

### E. Naïve Bayes Classifier

The naïve Bayes classifier (NBC) is perhaps the most superficial and widely studied supervised probabilistic ML method that uses Bayes' theorem with strong independence assumptions between the features to procure results. The NBC assumes that each input attribute variable is independent of training the data. This can be considered a naïve assumption about real-world data. Then, the conditional probability of each attribute $A_i$, given the class label $C$ is learnt from the training data (Duda and Hart, 1973).

The strength of the NBC lies in its ability to handle an arbitrary number of independent numerical or categorical attribute features. The solid but often controversial primary assumption (due to its "naivety") is that all the attributes $A_i$ are independent given the value of class $C$. For classification, the Bayes rule is applied to determine the class of the unknown instances by computing the probability of $C$ given $A_1, ..., A_n$ and then selecting the class with the highest posterior probability. The "naive" assumption of conditional independence of a collection of random variables is very important for the above result. Otherwise, it would be impossible to estimate all the parameters without such an assumption. This is a relatively strong assumption that is often not applicable. However, any bias in estimating probabilities may not make a difference in practice – it is the order of the probabilities and not the exact values that determine the probabilities.

Nonetheless, NBC has been shown to solve many complex real-world problems and to do so effectively. Also, it requires a small amount of training data to estimate the parameter. A frequency table is created for each attribute against the target (class) to calculate the posterior probability to classify an unknown instance. Then, the NBC is used to calculate the posterior distribution. Once again, the prediction outcome is the class with the highest posterior probability.

## 3. MULTIPLE CLASSIFIER LEARNING SYSTEMS

Multiple classifier learning systems (MCLS) can be defined as a set of classifiers whose individual predictions are combined in some way to classify new examples to produce one optimal predictive model. The most common type of MCLS includes an ensemble of classifiers to function for a parallel classifier input combination. However, a significant number of methods have been used to create and combine such individual classifiers, including ensemble methods, committee, classifier fusion, combination, aggregation, etc.

There are three types of MCLS architectures, namely - static parallel (SP); multi-stage (MS) design; and three dynamic classifier selection (DCS) [Twala, 2009; Finlay, 2011; Twala, 2016]

One of the most famous MCLS architectures is SP by Zhu *et al*. (2001). For SP, two or more classifiers are developed independently and executed in parallel; then, the outputs generated by all base classifiers are combined to determine a final classification decision (selected from a set of possible class labels). Many combination functions are available for this architecture, including majority voting, weighted majority voting, the product or sum of model outputs, the minimum rule, the maximum rule and Bayesian methods. Averaging is mainly used for regression problems, while voting is used for classification problems. There are two categories of SP-related MCLS: a single ML algorithm is used as base learning (homogenous parallel), and multiple ML algorithms are used as base learning (heterogeneous parallel). For the paper, the former category is used.

The second type of MCL system architecture is MS design, where the classifiers (usually with no overlaps) are organised into multiple groups and then iteratively constructed in stages. At each iteration, the parameter estimation process depends on the classification properties of the classifiers from previous stages. As with SP, this design benefits from processing inputs in parallel. It ensures that labels are assigned using only the necessary features. In addition, the number and composition of stages used by the model have proven to have a significant impact on overall performance. Some MS approaches have been used to generate models applied in parallel using the same combination rules used for SP methods. For example, most boosting strategies have been shown to create weak classifiers, but they tend to form stronger ones (Schapire *et al*., 1990).

A DCS is an ensemble learning architecture developed and applied to different regions within the problem domain. The technique involves training MCLS on the dataset and selecting the best prediction models. The *k*-NN approach is sometimes used to determine instances that are closely related to the unknown instance to be predicted (see Section 2). While one classifier may be shown to outperform all others based on global performance measures, it may not necessarily dominate all other classifiers entirely. Weaker competitors will sometimes beat the best across some regions (Kittler, 1997). Research has shown DCS performs better than single classifiers and even better than combining all the base classifiers. Furthermore, Kuncheva (2002) approached DCS problems from a global and local accuracy perspective with promising results.

Ensemble learning of classifiers can be classified into three stages: 1) generation, 2) selection, and 3) integration. The objective of the first stage is to obtain a pool of models, followed by a selection of a single classifier or a subset of the best classifiers. Finally, the base models are combined to obtain the prediction for new or unknown instances. The aspect of multiple classifier systems is determining the number of component classifiers in the final ensemble (also known as ensemble size or cardinality) is the most important. The impact of ensemble size on efficiency in time and memory and predictive performance makes its determination a critical problem (Kuncheva, 2002; Lin *et al*., 2004; Rokach, 2012; Hernandez-Lobato *et al*., 2013).

Furthermore, one should assume that diversity among component classifiers should be another influential factor in an accurate ensemble. However, no explanatory theory reveals how and why diversity among components contributes to overall ensemble accuracy. Therefore, all possible ensemble sizes and their respective diversities are considered for this paper.

Recently, Multi-Classifier-Based Boosting was introduced, where clustering and classifier training are performed jointly (Babanko et al., 2009; Kim and Cipolla, 2008). These methods have been applied to object detection, where the entire training set is available from the beginning. Other related works include multiple instance learning (Viola *et al.*, 2006; Jackowski, 2018) and multiple deep learning architectures (Mellema *et al.*, 2019). The former algorithm learns with bags of examples, which only need to contain at least one positive example in the positive case. Thus, the training data does not have to be aligned. Mellema et al. (2019) developed the system using anatomical and functional features to diagnose a subject as autistic or healthy.

## 4. EXPERIMENTS

### 4.1 Experimental set-up

When picking the right machine learning method for classification or predictive tasks one has to be decisive, with performance being the most dominant factor. The ensemble of classifiers has shown to achieve higher predictive performance in comparison with single classifiers and for many application tasks. In this paper, the underlying hypothesis is that the ensemble of classifiers will achieve higher ASD predictive performance or yield better results compared to single classifiers.

Our investigations are conducted using a dataset of behavioural data and robot-enhanced therapy recorded from 61 children diagnosed with ASD (Billing *et al.*, 2020). The dataset covers 3,000 therapy sessions and more than 300 hours of treatment. Half of the children interacted with the social robot supervised by a therapist, while the other half was used as a control group (i.e., interacting directly with the therapist). Both groups followed the Applied Behavior Analysis protocol. Each session was recorded with three red-green-blue (RGB) cameras and two Red-Green-Blue-Depth (RGBD) Kinect cameras, providing detailed information on children's behaviour during therapy; the dataset comprises body motion, head position and orientation, and eye gaze variables, all specified as 3D data in a joint frame of reference. In addition, metadata including participant age, gender, and autism diagnosis variables are included.

For the simulations, five base classifiers are modelled using default hyper-parameters for each respective classifier. Each approach utilises a different form of parametric estimation or learning. For example, they generate various forms in linear models, density estimation, trees, and networks. These classifiers are among the top 10 most influential and popular algorithms in data mining (Wu *et al.*, 2008). They are all practically applicable to ASD, with known examples of their application within the robotics-enhanced therapy industry.

First, each state-of-the-art classification method (base classifier) was constructed using MATrix LABoratory or MATLAB software (MATLAB, 2019). These base classifiers were later used and assessed as a benchmark against various MCLS. It was evident that the benefits of using ensembles could not be achieved by simply copying an individual model and combining the individual predictions. For this reason, all possible combinations of the number of classifiers per ensemble were explored (i.e., from two classifiers per ensemble to five classifiers). These ensembles are defined as Multiple Classifier Learning systems 2 (MCL 2) (for two classifiers per ensemble); Multiple Classifier Learning systems 3 (MCL 3) (for three classifiers per ensemble; Multiple Classifier Learning systems (MCL 4) (for four classifiers per ensemble), and Multiple Classifier Learning systems 5 (MCL 5) (for all five classifiers in the ensemble).

To assess the performance of the base classifiers, the training set -validation set-test set methodology is employed. First, the dataset was split randomly into a 60% training set for each run, a 30% validation set and a 10% testing set. To test the effectiveness of the classifiers, the dataset was further split randomly into 5-folds. The smoothed error rate (i.e., smoothing the normal error count using estimates of posterior probabilities and the posterior probabilities using Bayesian estimation with conjugate priors) was used as a performance measure in all the experiments. This rate was used primarily for its variance reduction benefit and for dealing effectively with a tie among two competing classes (Twala, 2005).

The DT classifier identifies and ranks the features (factors) that significantly impact or contribute to ASD. The set of features available forms the input to the algorithm with a DT as output. The purpose of these techniques was to discard irrelevant or redundant features (factors) from a given vector. For the paper, feature (factor) selection was used by evaluating the mutual information gain of each variable in the context of the target variable (detection or non-detection of ASD). Feature (factor) ranking and selection methods have been implemented with two basic steps of a general architecture for our experiments: subset generation and subset evaluation for the ranking of each feature in every dataset. Then, the filter method is used to evaluate each subset. Overall, a mutual information-based approach on the single classifier that exhibits the highest accuracy rate is utilised for this task. Mutual information calculates the reduction in entropy from the transformation of a dataset.

The fixed-effect model is used to test for statistical significance of the main effects (i.e. the five single classifiers; twenty-three multiple classifier systems, three multiple classifier architectures and five resampling procedures) versus their respective interactions. Each experiment is randomly replicated five times (5-fold) making it a total of 5 x 23 x 3 x 5 x 5 = 8625 experiments.

*4.2 Experimental results*

Experimental results on the Autism Spectrum Disorder predictive performance of single classifiers (on the one hand) and MCLS (on the other hand) are described. The behaviour of multiple classifiers is explored for different MCLS architectures and resampling procedures.

The results are presented in three parts.

The first part compares the performance and robustness of five SCLS in predicting Autism Spectrum Disorder in children. The second part investigates the performance of MCLS (i.e., ensembles, resampling procedures, and architectures) to determine if there is an improvement in ASD predictive accuracy. These overall results are for each MCLS. They are averaged for all ensemble learning combinations about resampling procedures and architectures. Then, the experimental comparison of MCLS (for all possible ensemble combinations) is presented. Finally, the behavioural factors (in ranking order) that contribute to and are critical when addressing the ASD problem have been identified.

Figures 1 to 9 plot the smoothed error of the instances learned on the target domain, averaged over five-fold cross-validation runs by each one of the methods. The same folds were used to evaluate each method. All the main effects (i.e., base or single classifier systems, MCLS, resampling procedures, and MCLS architectures) were significant at the 5% level, with F-ratios of 131.7, 71.4, 513.6 1132.6, respectively.

From Figure 1, it follows that DT is the best base classifier, exhibiting a smoothed error rate of 35.7% (or 0.643 accuracies). The second-best base classifier is ANN, followed by *k*-NN and LgD with smoothed error rates of 36.2% (0.638 accuracies), 38.5% (0.615 accuracies), and 41.7% (0.583 accuracies). Finally, the worst performance is by the NBC with a smoothed whooping error rate increase of 43.3% (0.567 accuracies). The most relevant attributes to predicting ASD are communication and interaction for the single classifiers.

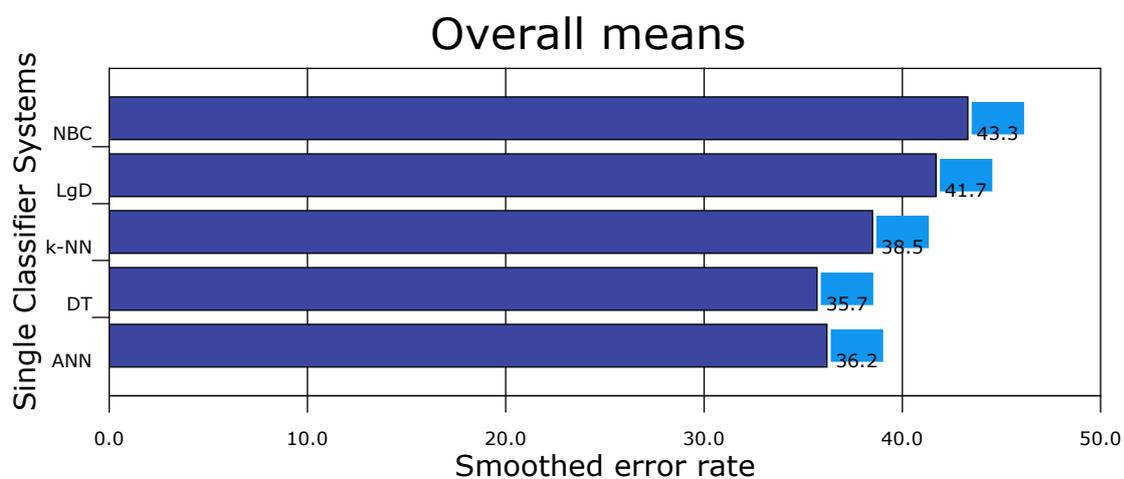

**Figure 1.** Single classifier systems

From Figure 2, it appears that the performance of all the MCLS is more significant when the ensemble size is composed of only three classifiers.- The smoothed error rate for this kind of ensemble is 21.4%. All ensembles with only two classifiers exhibit the second-best performance (a smoothed error rate of 30.7%), while those with only four classifiers take the third spot. The worst performance is when the ensemble comprises all five classifiers (with a smoothed error rate of 36.5%). The difference in performances between the four ensembles is statistically significant at the 5% significance level. Social communication and interaction were the most relevant features when predicting ASD (using multiple classifier systems).

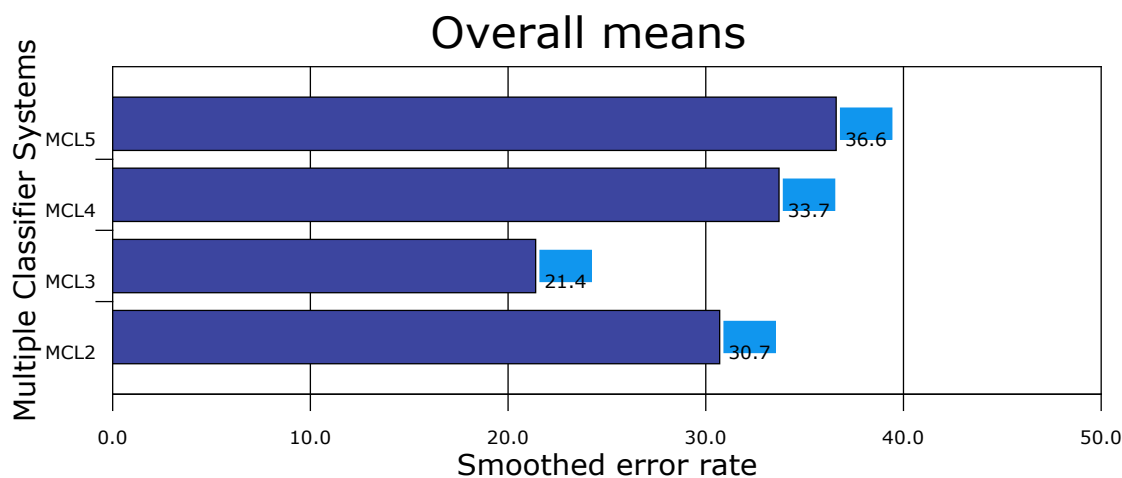

**Figure 2.** Multiple classifier systems

The results summarised in Figure 3 show multiple classifier systems achieving higher accuracy rates when bagging is used with a smoothed error rate of 0.233 (an accuracy rate of 76.7%), followed by boosting (a smoothed error rate of 0.259 or accuracy rate of 74.1%). The third best resampling procedure is feature selection (a smoothed error rate of 0.314 or an accuracy rate of 69.6%), followed by randomisation (a smoothed error rate of 0.347 or an accuracy rate of 65,3%). All multiple classifier systems perform poorly when staking is used as a resampling procedure (a smoothed error rate of 0.372 or an accuracy rate of 62.85). The experimental results of the three architectures used when constructing MCLS are related to the above.

From Figure 4, it appears that all the multiple classifier systems have a more significant robust effect when the multi-stage design is used as an architecture (accuracy rate of 73.5%), followed by static-parallel and dynamic classifier selection with accuracy rates of 71.8% and 67.5%, respectively. The difference in performance between the architectures was significant at the 5% level.

The results presented in Figure 5 shows all the MCLS performing worse under the dynamic classifier selection (an error rate of 32.5%) compared with single parallel and multi-stage. On the other hand, the MCLS performs slightly better when the multi-stage architecture design is used (26.5%) than a single parallel (28.2%). Thus, the difference in performance between the three architectures was significant at the 5% significance level (following a similar pattern to Figure 4 results).

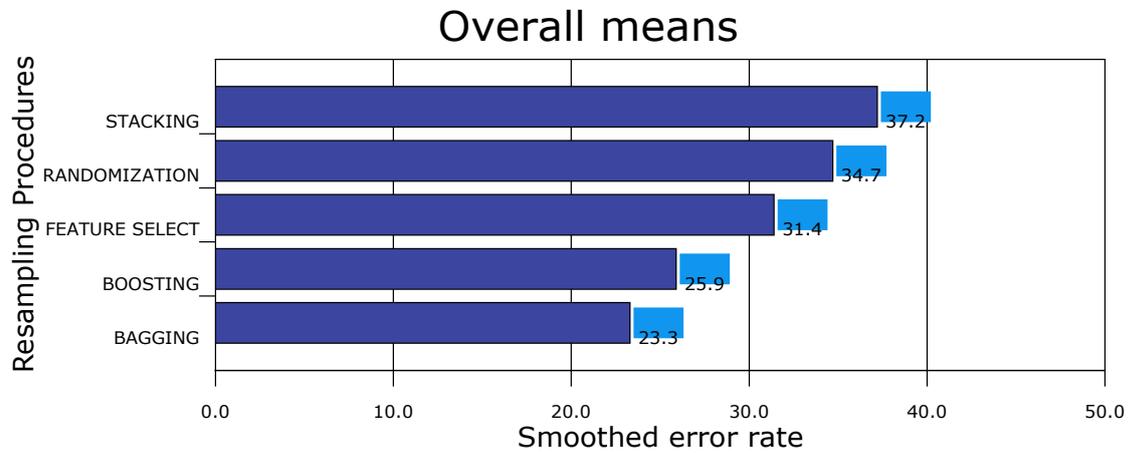

**Figure 3.** Resampling Procedures

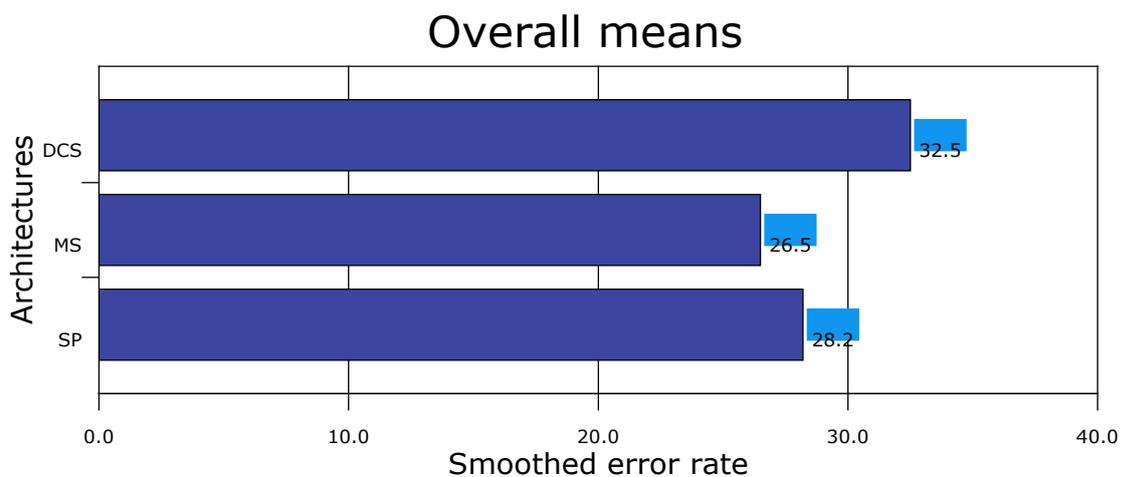

**Figure 4.** Multiple Classifier Systems Architectures

The results of the interaction effect (which was found to be statistically significant at the 5% level of significance) between MCLS, architectures and resampling procedures are displayed in Figure 5. It follows that all Multiple Classifier Systems perform differently from each other, with significant error rate increases observed for ensembles with five or four classifiers compared to those with three or two classifiers per ensemble. The difference in error rates between the two conditions is noticeable for multi-stage design and single parallel, with minor differences observed for dynamic classifier selection). Once again, the results show that MCLS built through bagging is the best technique for predicting ASD, followed by boosting, feature selection, randomisation and stacking, respectively.

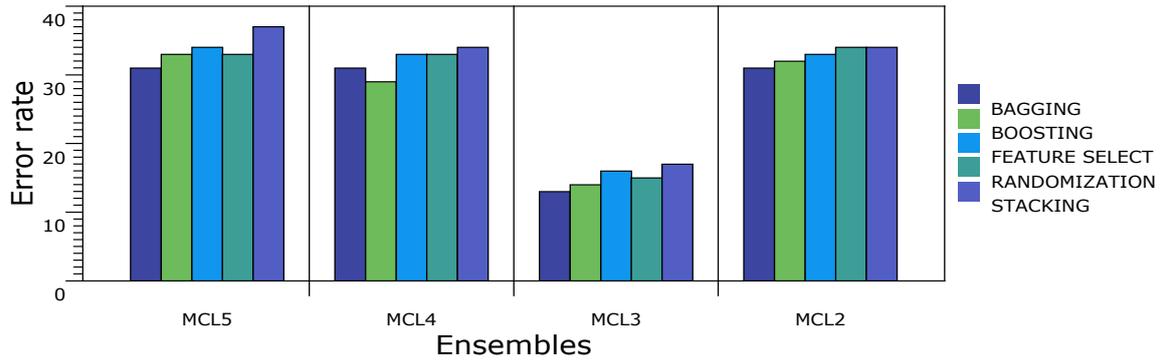

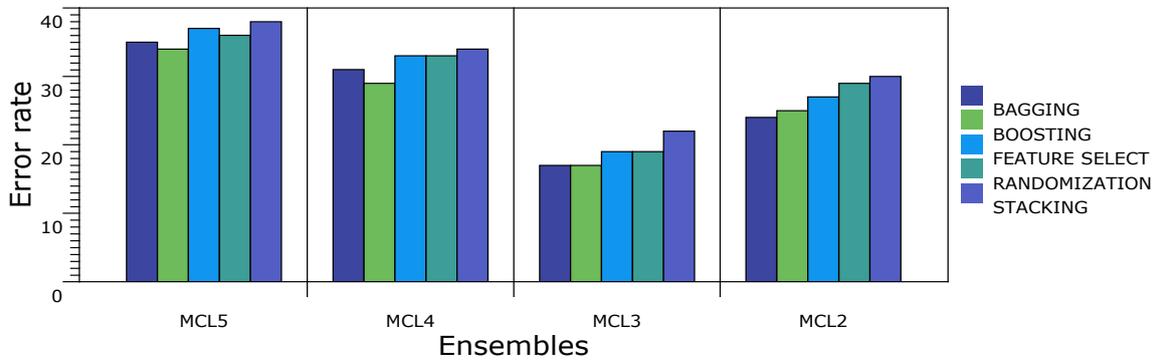

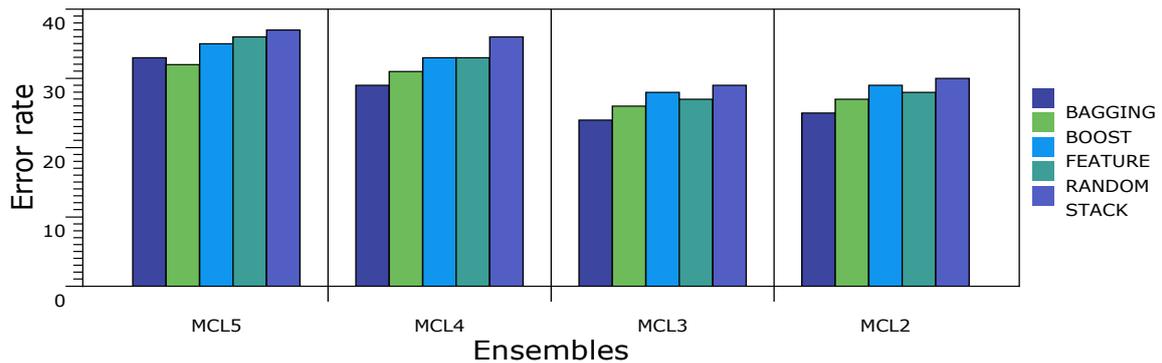

**Figure 5.** Multiple Classifier Learning Systems (Overall Results)

From Figure 6A, the effect of the resampling procedures Multiple Classifier Learning System 2 (MCLS 2) is transparent. MCLS 2 exhibits the worst performance for stacking, closely followed by feature selection and randomisation. The best overall performance for a static parallel architecture comes about when bagging is used. In contrast, the best performance is observed when decision trees and logistic discrimination are the two components of the ensemble. The ensembles of Artificial Neural Network and Decision Trees and Logistic Discrimination and naïve Bayes classifier exhibit the worst performances.

From Figure 6B, bagging exhibits minor error rate increases (with tight competition from boosting) for MCLS 2 and when the multi-stage architecture is used. One striking outcome is the Artificial Neural Networks and Logistic Discrimination ensemble performance, which compares favourably with a Decision Tree and Logistic Discrimination ensemble. However, the ensembles with Artificial Neural Network and Decision Trees and Logistic Discrimination and naïve Bayes classifiers exhibit one of the worst performances for MCLS 2. Another poor performance is when the *k*-nearest neighbour and naïve Bayes classifiers are used as ensemble components, primarily when randomisation is used.

The dynamic classifier selection system is observed. At the same time, stacking continues to struggle and achieves the worst performance, especially when the Artificial Neural Network and a Decision Tree (on the one hand) and logistic discrimination and naïve Bayes classifier (on the other hand) are components of the ensemble (Figure 6.9C). The best performing ensembles are Artificial Neural Network and logistic discrimination, and the Decision Tree and Logistic Discrimination are components.

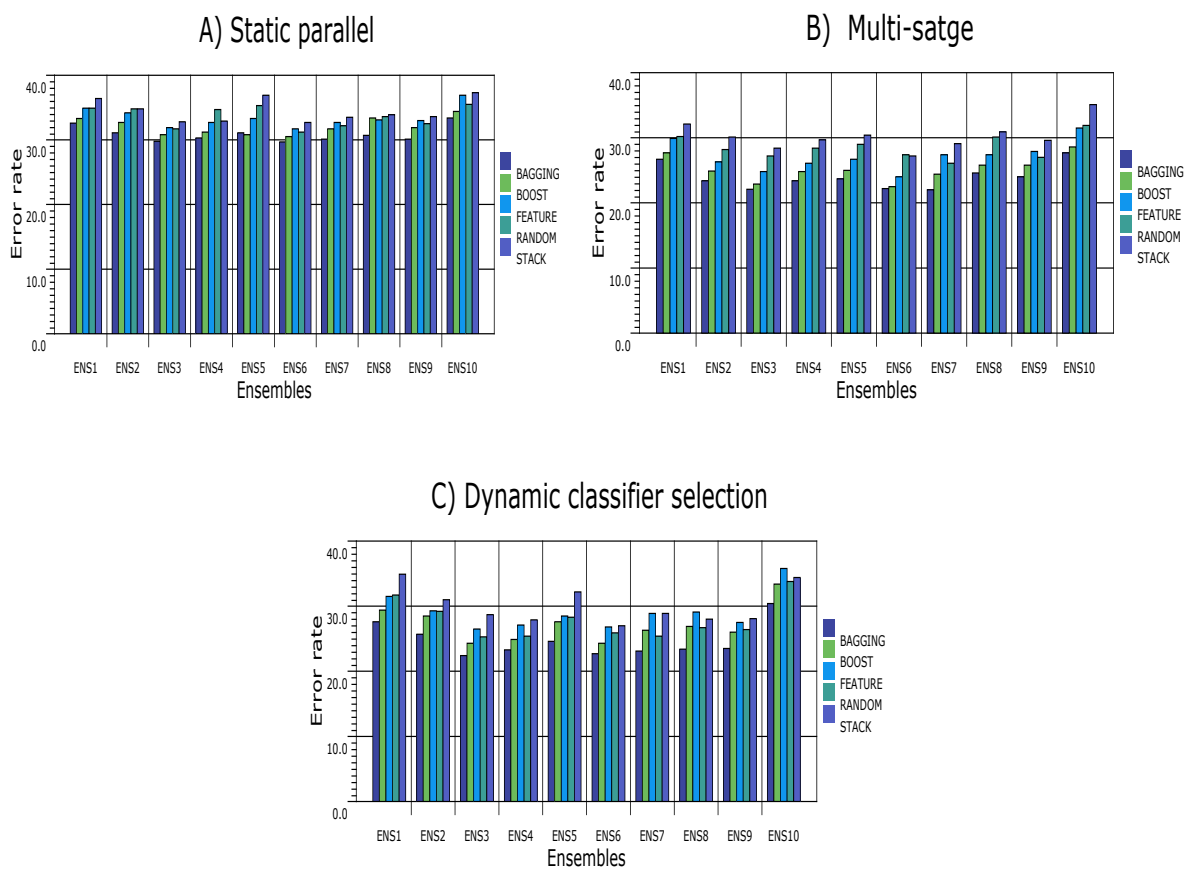

**Figure 6.** Multiple classifier learning systems 2

The performance of methods for Multiple Classifier Learning (MCL3) follows a similar pattern to the one observed for MCL2 (Figure 7).

Figure 7A also shows smaller increases in error rates for all the resampling procedures for static parallel than the same architecture for MCL2. The best performing ensemble is when Decision Trees, *k*-nearest neighbour and logistic discrimination are components. On the other hand, poor performances are observed when the Artificial Neural Network, Decision Trees and k-nearest neighbour and Artificial Neural Network, and *k*-nearest neighbour and naïve Bayes classifiers are components of the ensemble. This is the case for the feature selection and stacking resampling procedures.

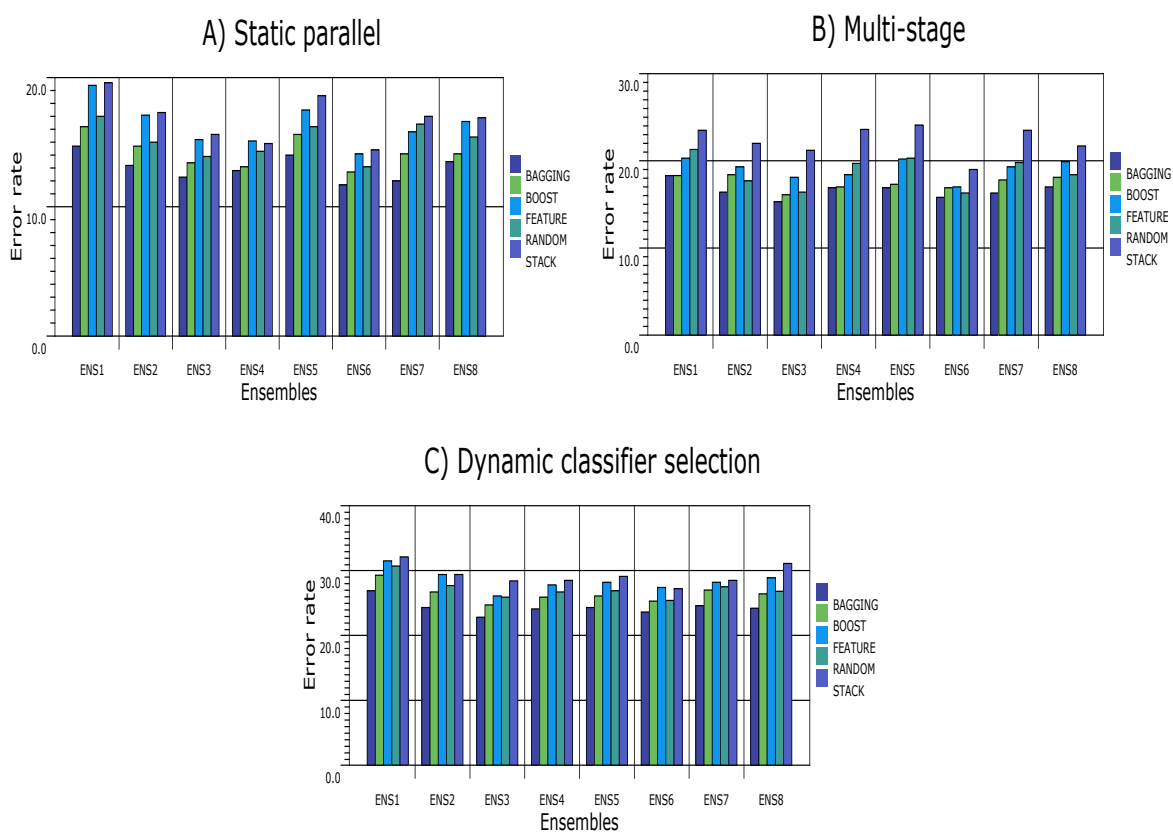

**Figure 7.** Multiple classifier learning systems 3

The methods for multi-stage design (Figure 7B) are nearly identical to those observed for MCLS2, with all ensembles achieving higher accuracy rates when bagging and boosting are used. Otherwise, on average, the performance of all the methods worsens when stacking is used. The best performing ensemble is the decision tree, *k*-nearest neighbour, and Logistic Discrimination (primarily feature selection, randomisation and stacking). For stacking, the ensemble method composed of an Artificial Neural Network, *k*-nearest neighbour, and naïve Bayes classifiers proves to be the worst-performing.

The impact of MCLS 3 on predictive accuracy is shown in Figure 7C. Once again, bagging yields the best performance, closely followed by boosting with severe competition from randomisation. Once again, the best performing ensemble is when the Artificial Neural Network, Decision Trees and naïve Bayes classifiers are components. The ensemble with the *k*-nearest neighbour, logistic discrimination and naïve Bayes classifier drops from being the third-best performing (when stacking and multi-stage design is used) to be one of the worst (when stacking and dynamic classifier election are used).

Overall all the MCLS 3 systems perform better when static parallel is used, followed by multi-stage and dynamic classifier selection, respectively.

Figure 8A follows that when using static parallel to build an MCLS 4, boosting is the best technique for dealing with the autism spectrum disorder problem, with an Artificial Neural Network, *k*-nearest neighbour, logistic discrimination, and naïve Bayes classifiers as components for the ensemble. On the other hand, the ensemble with an Artificial Neural Network, Decision Trees, *k*-nearest neighbour and logistic discrimination as components achieves the worst performance. This is the case at all resampling procedure levels (i.e. bagging, boosting, feature selection, randomisation and stacking).

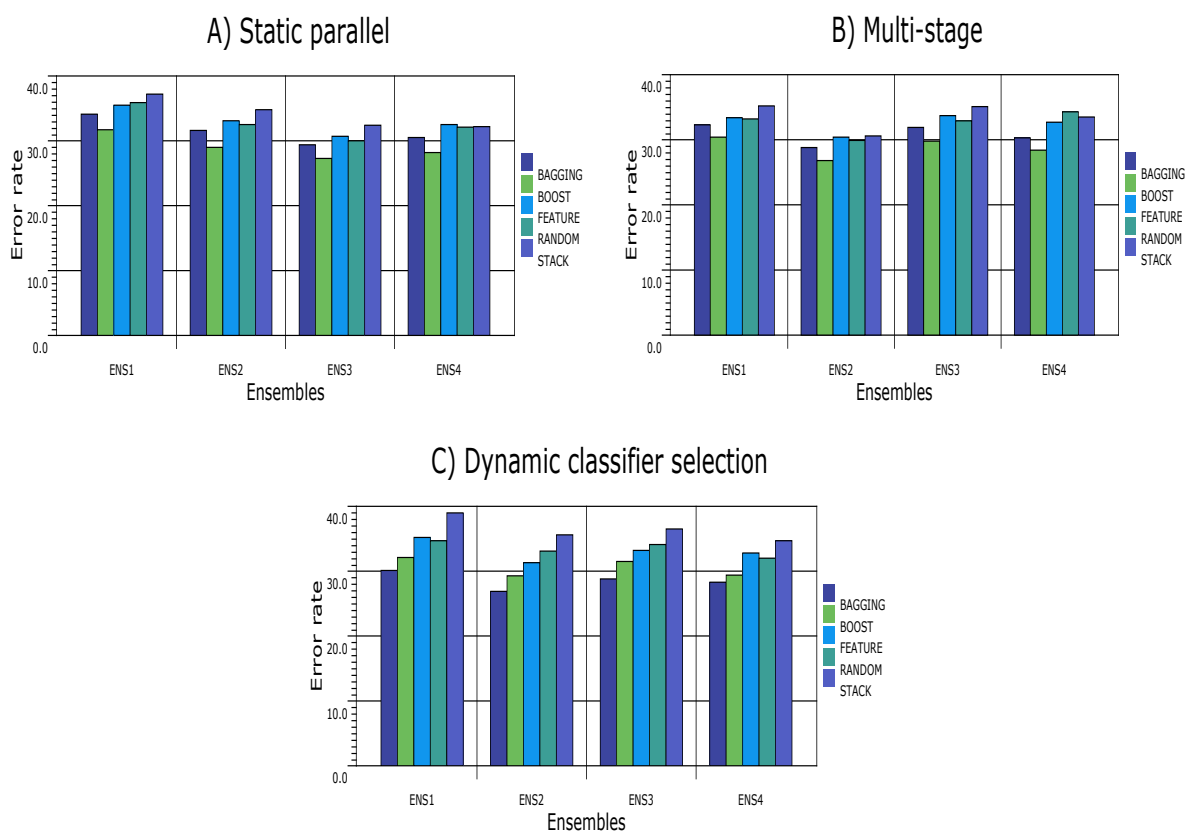

**Figure 8.** Multiple Classifier Learning System 4

It follows from Figure 8B that the best technique for handling Autism Spectrum Disorder for a multi-stage design and across various resampling procedures is boosting, closely followed by bagging. However, poor performances are observed for feature selection, randomisation, and stacking methods. Also, the ensemble with an Artificial Neural Network, Decision trees, *k*-nearest neighbour and logistic 0 discrimination as components exhibits the worst performance.

For MCLS 4, bagging using dynamic classifier selection shows superior performances to the other resampling procedures (Figure 8C). The best performing ensemble (across bagging, boosting, and feature selection) is where components of Artificial Neural Network, Decision Trees, k-nearest neighbour and naïve Bayes. On the other hand, randomisation and stacking of an ensemble with an Artificial Neural Network, k-nearest neighbour, logistic discrimination, and naïve Bayes perform best.

Once again, all the ensembles appear to perform better when the multi-stage design is used, followed by static parallel and dynamic classifier selection, respectively.

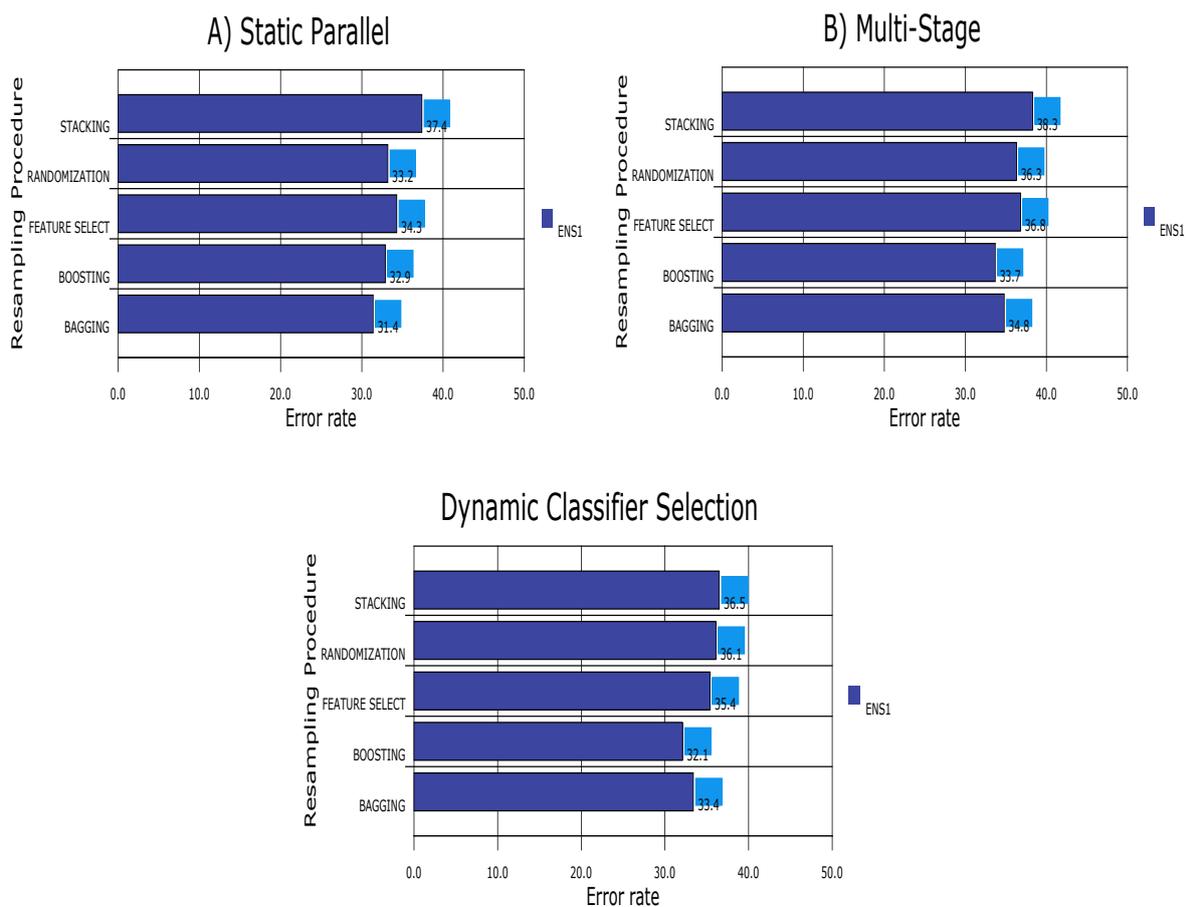

**Figure 9.** Multiple classifier learning systems 5

For this kind of problem, it seems that building an MCLS 5 using static parallel performs better compared with methods like dynamic classifier selection and multi-stage design (Figure 9). Also, boosting appears to be more effective than usual, mainly when dynamic classifier selection and multi-stage design are used, outperforming methods like feature selection, randomisation and stacking in some situations. Another good performance is randomisation when missing static parallel, and a multi-stage design is used. On the other hand, stacking is the worst-performing method across all the architectures.

The goal of feature selection in ML is to find the best features to build applicable models of a studied phenomenon (for example, removing non-informative or redundant predictors from the model). Many feature selection techniques are classified into supervised (wrapper filter, intrinsic, embedded) and unsupervised learning (for unlabelled data). The Decision Tree algorithm (a supervised learning and embedded approach) was used to select the features in ranking order and according to the mutual information criterion whereby node impurities in the decision tree are utilised. The feature selection process algorithm results modelled to obtain features considered most relevant to ASD and their merit value ranks each feature are analysed in Table 1.

TABLE I. FEATURES OF ASD SORTED BY RELEVANCE

| **Features** | **Cross-validation error** |
|---|---|
| Communication | 19.43 ± 0.12 |
| Social Interaction | 13.33 ± 0.34 |
| Module | 22.86 ±0.19 |
| Play | 16.07 ± 0.28 |
| Social Communication (gestures) | **7.51 ± 0.14** |
| Stereotype | 24.52 ± 1.45 |

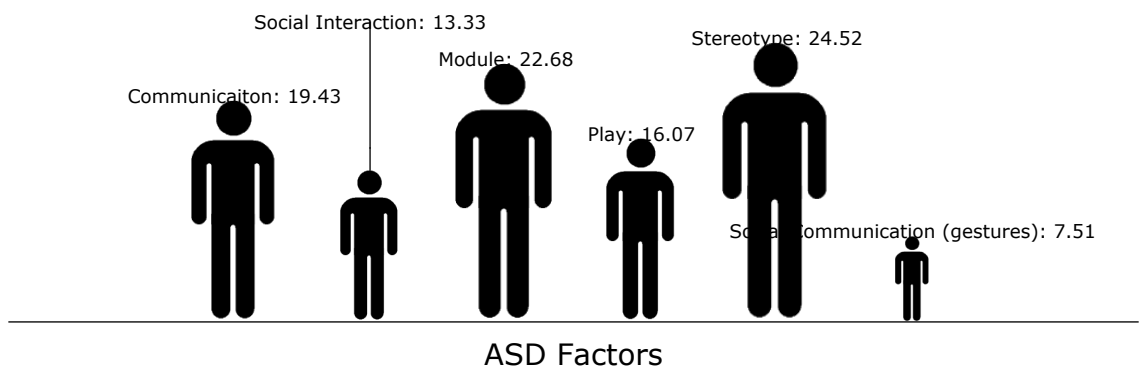

**Figure 10.** ASD factors

In terms of ranking, the results show *social communication* yielding the slightest cross-validation error (7.51%) followed by *social interaction* (13.33%) and *play* (16.07%), respectively. Otherwise, *stereotype* and *module* are the two featured exhibiting error rates of more than 20%, i.e., 24.52% and 22.86%, respectively. In addition, all the features were significantly different at the 5% level of significance.

## 5. REMARKS AND CONCLUSION

This paper aims to investigate the predictive performance of an ensemble of classifiers utilising psychometric measures, clinical observation, genetic, neurobiological, and physiological research from children (ages 3 to 6) using features to predict Autism Spectrum Disorder.

Open questions related to predicting with confidence addressed include how can data be utilised effectively to achieve more efficient confidence-based predictions using ensemble classifiers. To this end, the significant contributions of the paper include showing the robustness of single classifiers for predicting Autism Spectrum Disorder and showing how MCLS provide improvements in performance over the SCLS (including the best performing one). Finally, the best features (according to mutual information-based ranking) that influence Autism Spectrum Disorder are identified.

The results show dynamic classifier selection that segments the population in several sub-regions as consistently performing poorly. All the experiments yield results that are inferior to the single best classifier. However, the performance of most static parallel and multi-stage combination strategies provides statistically significant improvements over the single best classifier. Ensembles with a combination of three classifiers outperform the other MCLS, with stacking achieving a poor performance compared to other resampling procedures such as boosting, feature selection, and randomisation. The most exciting result is bagging performance, which consistently outperforms all other multiple classifier systems, especially for ensembles with a combination of three single classifiers.

Furthermore, social communication appears to be the critical behavioural factor to be considered when dealing with ASD. This can be argued because of the inability of children with this disorder to communicate and use language, which depends heavily on their intellectual and social development. In other words, some children with ASD may not be able to communicate using speech or language, and some may have minimal speaking skills. Therefore, social interaction is another factor that needs consideration when dealing with ASD.

The conclusions are that single training classifiers can obtain influential ensembles in several different ways. Still, that high average individual accuracy or much diversity would generate influential ensembles. Several findings regarding diversity and compelling ensembles presented in the literature in recent years are also discussed and related to the results of the included studies. When creating confidence-based predictors using conformal prediction, several open questions regarding how data should be utilised effectively when using ensemble learning. Nonetheless, the paper has shown how accurately predicting ASD symptom severity would allow clinicians to confidently diagnose and assign the most appropriate intervention. Hence, saving countries a lot of money.

The study is based on children over the age of three years old. In a subsequent study, the datasets of children of all ages will be critically analysed to train the prediction model. The focus will be to collect more data from various sources and age groups and improve the proposed ML classifier to enhance its accuracy. A more state-of-the-art classification method will also be considered.

In sum, this research provides an effective and efficient approach to predicting and detecting autism traits for children above three years. This is because tests and diagnoses of autism traits are costly and lengthy. The difficulty of detecting autism in children and adolescents does not help another cause for the delay in diagnosis. Thus, with the help of accurate autism spectrum disorder predictive accuracy, an individual can be guided early to prevent the situation from getting any worse and reduce costs associated with such delay.

**ACKNOWLEDGEMENTS**

Thanks are extended to Swedish Data National Service for making the dataset available. There are no biomedical financial conflicts of interest to disclose. Also, this research received no specific grant from any funding agency in the public, commercial, or not-for-profit sectors.

**REFERENCES**

[1] Allison, C., Auyeung, B. & Baron-Cohen, S. (2012). Toward brief "red flags" for autism screening: the short autism spectrum quotient and the short quantitative checklist in 1 000 cases and 3 000 controls, *Journal of the American Academy of Child & Adolescent Psychiatry*, 51, pp.

[2] Aha, DW, Kibbler, DW & Albert, M.K. (1991). Instance-based learning algorithms. *Machine Learning*, 6 (37): pp. 37-66.

[3] Bone, B., Bishop, SL., Black, MP., Goodwin, MS., Lord C., & Narayanan, SS. (2016). Use of machine learning to improve autism screening and diagnostic instruments: effectiveness efficiency and multi-instrument fusion, *Journal of Child Psychology and Psychiatry*, 57. pp

[4] Breiman, L. (1996). Bagging predictors. *Machine Learning*, 26 (2): pp. 123-140.

[5] Breiman, L., Friedman, J., Olshen, R. & Stone, C. (1984). *Classification and Regression Trees*. Wadsworth.

[6] Buescher, AVS., Cidav, Z., Knapp, M. & Mandell, DS (2014). Costs of autism

spectrum disorder in the United Kingdom and the United States, *Journal of the American Medical Association Paediatrics*, 168 (8): pp. 721-728.

[7]   Cox, DR. (1966) Some procedures associated with the logistic qualitative response curve. In *Research Papers in Statistics: Festschrift for J. Neyman* (ed. F.N. David), Wiley, New York, pp. 55-71.

[8]   Dietterich, T. (2000). An experimental comparison of three methods for constructing ensembles of decision trees: bagging, boosting, and randomisation. *Machine Learning*, 40 (2): pp. 139-158.

[9]   Duda, RO. and Hart, PE. (1973). *Pattern Classification*, 2nd Edition, New York: John Wiley & Sons.

[10]  Gilat, A. (2004). *MATLAB*: *An Introduction with Applications*. 2nd Edition. John Wiley & Sons.

[11]  Finlay, S. (2011). Multiple classifier architectures and their application to credit risk assessment, *European Journal of Operational Research*, 210 (2): pp. 368-378.

[12]  Freund, Y. & Schapire, R. (1996). A decision-theoretic generalisation of online learning and an application to boosting. *Journal of Computing and Systems*, 55: 119-139.

[13]  Heinsfeld, AS, Franco, AR., Craddock RC, Buchweitz, A., & Meneguzzi, F. (2018). Identification of autism spectrum disorder using deep learning and the abide dataset. *NeuroImage: Clinical*, 17.

[14]  Hernandez-Lobato, D., Martınez-Munoz, G. & Suarez, A. (2013). How large should ensembles of classifiers be? *Pattern Recognition*, 46 (5): pp. 1323–1336.

[15]  Ho, TK. (1995). Random decision forests. In *Proceedings of the 3rd international conference on document analysis and recognition*, pp. 278-282.

[16]  Hong, S-K, Vodelstein, JT., Gozzi, A., Bernhardy, BC., Yeo, BTT., Milham, MP. & Di Martino, A. (2020). Toward Neuro Subtypes in Autism. *Biological Psychiatry*, 88 (1), pp. 111.128.

[17]  Hosmer, DW. & Lameshow, S. (1989). Applied Logistic Regression. Wiley: New York.

[18]  Jackowski, K. (2018). New diversity measure for data stream classification ensembles. *Engineering Applications of Artificial Intelligence*, 74: pp. 23–34.

[19]  Jolliffe, I. (1986). Principal Component Analysis. Springer Verlag.

[20]  Kirk, EE. (1982). *Experimental design* (2nd Ed.). Monterey, CA: Brooks, Cole Publishing Company.

[21]  Kittler, J., Hatef, M., Duin, R.P.W. & Matas, J. (1998). By combining classifiers. *IEEE Transaction on Pattern Analysis and Machine Intelligence*, 20 (3): pp. 226-239.

[22]  Klingspor, V., Morik, K. & Rieger, A. (1995). Learning Concepts from Sensor Data of a Mobile Robot. *Machine Learning-Special issue on robot learning*, 23 (2-3): pp. 305-332.

[23]  Kononenko, I. (1991). Semi-naïve Bayesian classifier. In *Proceedings of the European Conference on Artificial Intelligence*, pp. 206-219.

[24]  Kuncheva, L.I. (2002). Switching between Selection and Fusion in Combining Classifiers: An Experiment. *IEEE Transactions on Systems, Man and Cybernetics-Part B: Cybernetics*; 32.

[25]  Leroy, G. Irmscher, A. &Charlop-Christy, M.H. (2006) Data Mining Techniques

to Study Therapy Success with Autistic Children. In *International Conference on Data Mining*, 26 -29 June 2006, Monte Carlo Resort, Las Vegas, USA.

[26] MATLAB. (2019). *version 9.6 (R2019a)*. Natick, Massachusetts: The MathWorks Inc.

[27] Mellema, C., Treacher, A., Nguyen, K., & Montillo, A. (2019). Multiple Deep Learning Architectures Achieve Superior Performance Diagnosing Autism Spectrum Disorder Using Features Previously Extracted From Structural And Functional Mri. *IEEE 16th International Symposium on Biomedical Imaging (ISBI 2019)*, pp. 1891-1895.

[28] Quinlan, JR. (1993). *C.4.5: Programs for Machine Learning*. Los Altos, California: Morgan Kauffman Publishers, Inc.

[29] Ripley, BD. (1992). *Pattern Recognition and Neural Networks*. Cambridge University Press, New York: John Wiley.

[30] Rokach, L. (2010). Ensemble-based classifiers. *Artificial Intelligence Review*, 33 (1): pp. 1–39.

[31] Rumelhart, DE., Hinton, G. E., & Williams, R. J. (1986). Learning internal representations by error propagation. In Rumelhart, D. E. and McClelland, J. L., editors, *Parallel Distributed Processing*, 1: pp. 318-362. MIT Press.

[32] Safavian, SR. & Landgrebe, D., 1991. A Survey of Decision Tree Classifier Methodology. *IEEE Trans. Syst. Man Cybernet.* **21:** pp. 660–674.

[33] Schapire, R., Freund, Y., Bartlett, P. & Lee, W., 1997. Boosting the margin: a new explanation for the effectiveness of voting methods. *Proceedings of International Conference on Machine Learning*, Morgan Kaufmann, San Francisco: pp. 322–330.

[34] Soul, JS and Spence, SJ. (2020). Predicting Autism Spectrum Disorder in Very Preterm Infants. *Paediatrics*. 146 (4) e2020019448; DOI: https://doi.org/10.1542/peds.2020-01944

[35] Sowetan LIVE (2022). *Children with autism are excluded from school* [Accessed April 2022]. [Online]. Available: https://www.sowetanlive.co.za/news/south-africa/2022-04-05-children-with-autism-excluded-from-schools/

[36] Twala, B. (2016). Toward Accurate Software Effort Prediction using Multiple Classifier Systems. In Pedrycz, W and Succi, G and Silliti, A. (Eds.), *Computational Intelligence and Quantitative Software Engineering,* pp 135-151. Springer-Verlag.

[37] Twala, B. (2009). Multiple Classifier Learning to Credit Risk Assessment. *Expert Systems and Applications*, 37 (2010): pp. 3326-3336.

[38] Twala, B. (2005). *Effective Techniques for Dealing with Incomplete Data when Using Decision Trees*. Published PhD thesis, Open University, Milton Keynes, UK.

[39] van den Bekerom, B. (2017). Using machine learning for the detection of autism spectrum disorder

[40] Viola, P. Platt, J. C. & Zhang, C. (2006). *Multiple instances boosting for object detection*. In NIPS, pp. 1417–1426.

[41] Wall, D., Kosmicki, J., Deluca, T., Hastard, E., & Fusaro, V. (2012). Use of machine learning to shorten observation-based screening and diagnosis of autism. *Translational psychiatry*, 2 (4), pp. e100.

[42] Wolff, JJ. & Piven, J (2020). Predicting Autism in Infancy. *Journal of American Academy of Child & Adolescent Psychiatry*, S0890-8567(20)32061-X. DOI: 10.1016/j.jaac.2020.07.910.

[43] Wolpert, D. (1992). Stacked generalisation. *Neural Networks*, 5 (2): pp. 241-259.

[44] World Health Organization. (2018). *International classification of diseases for mortality and morbidity statistics* (11th Revision).

[45] World Health Organization (2017). *Autism spectrum disorders* [Accessed August 22, 2020]. [Online]. Available: http://www.who.int/news-room/fact-sheets/detail/autism-spectrum-disorder

[46] Wu, X., Kumar, V., Ross Quinlan, J. *et al.* (2008). Top 10 algorithms in data mining. *Knowl Inf Syst* 14, pp. 1–37.

[47] Xu, L., Geng, X., He, X., Li, J. & Yu, J. (2019). Prediction in autism by deep learning short-time spontaneous hemodynamic fluctuations. Front. Neurosci., 13, pp. 1120.

[48] Zhu, H., Beling, P.A., & Overstreet, G.A. A study in the combination of two consumer credit scores. *Journal of the Operational Research Society*, 52: pp. 2543-2559.